\documentclass[letterpaper]{article} 
\usepackage{aaai2026}  
\nocopyright
\usepackage{times}  
\usepackage{helvet}  
\usepackage{courier}  
\usepackage[hyphens]{url}  
\usepackage{graphicx} 
\usepackage{natbib}  
\usepackage{caption} 
\usepackage{algorithm}
\usepackage{algorithmic}

\usepackage{amsmath}
\usepackage{graphicx}
\usepackage{textcomp}
\usepackage{xcolor}
\usepackage{bm}
\usepackage{url}
\usepackage{multirow}
\usepackage{csquotes}
\usepackage{tabularx}
\usepackage{subcaption}
\usepackage{booktabs}
\usepackage{dsfont}
\usepackage{comment}
\usepackage{newfloat}
\usepackage{listings}
\DeclareCaptionStyle{ruled}{labelfont=normalfont,labelsep=colon,strut=off} 
\floatstyle{ruled}
\newfloat{listing}{tb}{lst}{}
\floatname{listing}{Listing}
\title{FMMO: Detecting the Divergence Between Local Attribution and Global Drift}
\author {
    Muhammad Rehman Zafar\textsuperscript{\rm 1},
    Ali El-Sharif\textsuperscript{\rm 2},
    Naimul Khan\textsuperscript{\rm 1}
}
\affiliations {
    \textsuperscript{\rm 1}Toronto Metropolitan University, Toronto, Ontario, Canada\\
    \textsuperscript{\rm 2}St. Clair College, Windsor, Ontario, Canada\\
    muhammadrehman.zafar@torontomu.ca, aelsharif@stclaircollege.ca, n77khan@torontomu.ca
}

\begin{document}

\maketitle

\begin{abstract}
Post-deployment drift poses a critical risk to algorithmic accountability, particularly when ground truth labels are delayed and performance degradation becomes a ``silent failure''. While Explainable AI (XAI) is often relied upon to audit these shifts, we demonstrate that popular local attribution methods (e.g., TreeSHAP) can exhibit misleading stability even as model reliability collapses. In this paper, we propose a \emph{Framework for Model Monitoring and Observability (FMMO)} designed to expose the divergence between local explanation stability and global distribution shifts. Using benchmark, synthetic, and real-world datasets, we show that local XAI methods fail to flag drift-induced disparate impact, specifically where False Positive Rates spike for protected groups while feature attributions remain unchanged. By integrating global surrogate models with model utilization measurements, FMMO mitigates this fairness blind spot, ensuring that stakeholders can detect discriminatory deterioration that standard local XAI tools overlook.
\end{abstract}


\section{Introduction}
Machine Learning (ML) models are trained to avoid overfitting training data and generalize to new unseen data~\cite{hastie2009esl}. The model's ability to generalize is contingent on the assumption that new unseen data will exhibit characteristics similar to those of the training data. Furthermore, once ML models are trained and deployed, their logic is fixed. ML models deployed in dynamic environments often encounter shifts in the distributions of inference data~\cite{biaek2025estimating,wu2021labelshift,zhang2013domain} known as data drift~\cite{nigenda2022amazon}, which can cause unobservable drops in predictive accuracy~\cite{mallick2022matchmaker}. Data drift refers to a gradual or sudden shift between historical training data and future test data, which can cause a significant drop in performance and reduce the overall efficiency of the system~\cite{mallick2022matchmaker}. This problem typically occurs when the ML model is used to make predictions on test data that differ from the original data on which it was trained. 

Therefore, model monitoring and observability are necessary to identify potential degradation in model performance during inference. This is particularly important in high-stakes domains, such as ML models deployed in finance, justice, or clinical settings~\cite{subasri2025harmful}. Model monitoring and observability can be challenging in scenarios where the ground truth is unavailable or not immediately available, and model performance cannot be evaluated by computing accuracy metrics comparing model predictions with ground truth \cite{biaek2025estimating,zliobaite2010change}. These ML models are at risk of silent failures in which performance degradation cannot be directly observable during inference. To address these limitations, proxy methods can be used to estimate model performance by leveraging access to observable inputs, model usage, and outputs. Some options for model monitoring during inference include feature monitoring (observe changes in the statistical distribution of input)~\cite{muller2024open} and model utilization monitoring (observe changes in feature importance during inference, derived from post-hoc Explainable Artificial Intelligence (XAI) methods, or by measuring the average confidence in inference predictions)~\cite{mougan2025explanation,rawal2025evaluating}.

\begin{figure}[ht]
    \centering
    \begin{subfigure}{0.45\linewidth}
        \centering
        \includegraphics[scale=0.30]{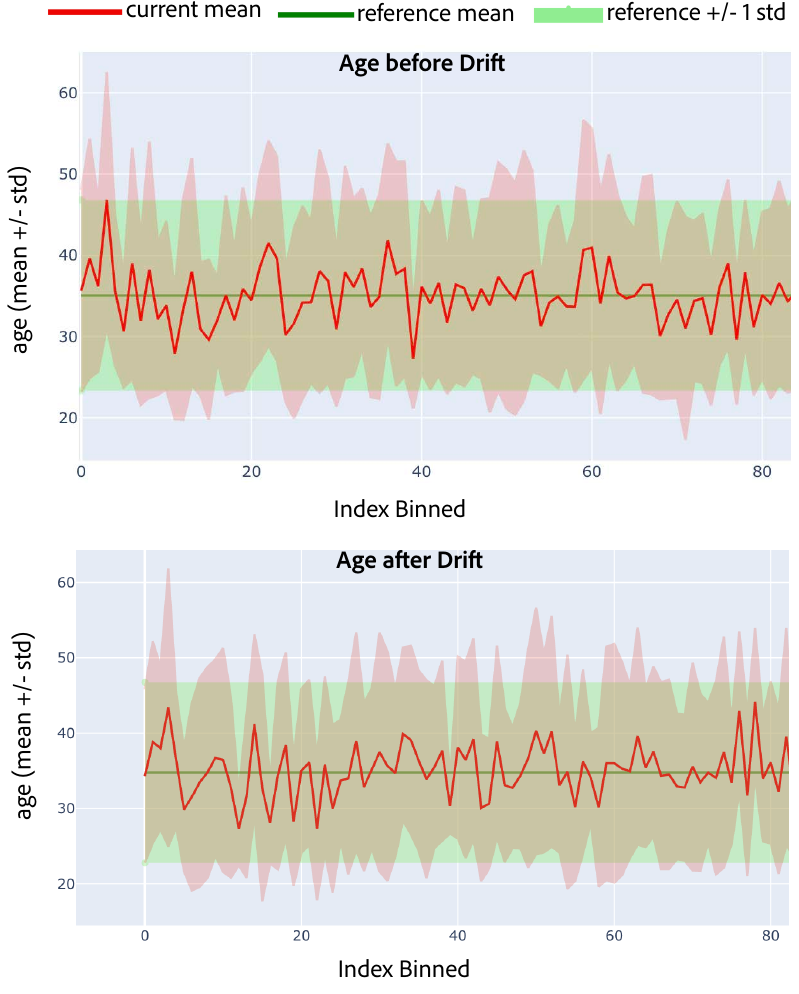}
        \caption{Subtle drift in the \enquote{Age} attribute was not detected by the univariate drift detector. }
        \label{fig:age_no_drift_detected}
    \end{subfigure}
    \hspace{1em}
    \begin{subfigure}{0.45\linewidth}
        \centering
        \includegraphics[scale=0.30]{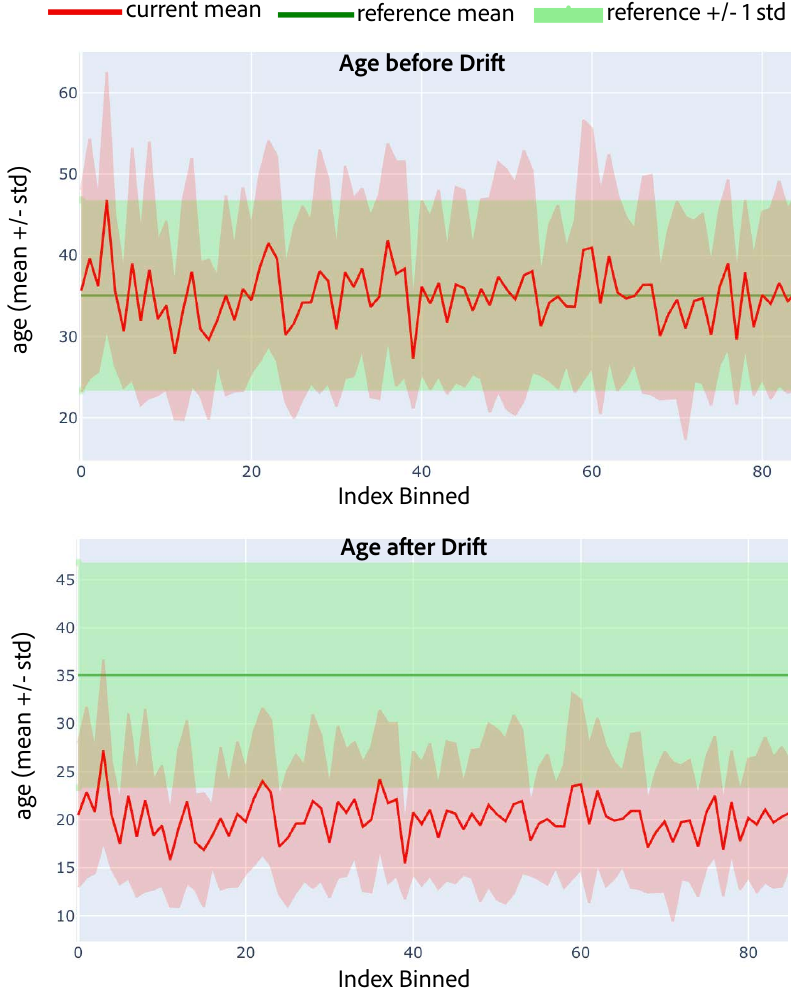}
    \caption{An example of Age column from the COMPAS dataset before and after drift.}
    \label{fig:drift_example}
    \end{subfigure}
    
    \caption{FMMO reports for COMPAS dataset when Gaussian noise was added.}
    \label{fig:drfit_non_drift}
\end{figure}


All observability techniques exhibit certain limitations. For instance, the univariate drift detector failed to identify subtle drift in attribute \enquote{Age} (Figure~\ref{fig:age_no_drift_detected}), although it affected model behavior by increasing the False Positive Rate (FPR) under low drift magnitudes (Table~\ref{tab:compas_rf_group_metrics}). A key limitation of the Domain Classifier (DC) is that, as a classifier-based drift detector, it only signals the presence of drift without providing insight into which features or mechanisms have changed.  As a result, DC offers limited transparency and does not explain how the drift is reflected in the data distribution. In addition, Figures~\ref{fig:dt_shap_baseline_ces} and \ref{fig:dt_shap_drift_ces} illustrate the constraints of local XAI methods.  A key limitation of Confidence-Based Performance Estimation (CBPE) is that it indirectly infers performance degradation from the predicted confidence scores of the model rather than from the true labels. As a result, CBPE can indicate that performance is deteriorating but cannot explain why the degradation occurs or which features or groups are responsible.  
Each auditing module targets a specific aspect of model monitoring under drift and is inadequate alone. Together, these techniques enable complementary drift detection, performance tracking, and explanation, resulting in a more robust and transparent auditing framework that connects distributional shifts to performance loss.

Therefore, we propose using multiple methods together to improve the probability of detecting a degradation in predictive accuracy. In this work, we present a robust Framework for Model Monitoring and Observability (FMMO) for flagging models at risk of performance degradation due to covariate drift. The FMMO integrates local and global XAI methods with model usage monitoring so that if any method detects performance degradation, it can serve as a potential indicator of covariate drift, triggering an alert for ML model auditing. The FMMO has direct implications for the governance of deployed ML models, particularly in settings where ground-truth labels are delayed, incomplete, or unavailable. By jointly tracking confidence-based performance estimation and explanation stability, the framework enables ongoing accountability for model behavior beyond initial validation, supporting continuous monitoring and observability of model performance during inference. Our contributions can be summarized as follows:
\begin{itemize}
    \item We propose a framework that jointly evaluates model performance and covariate drift along with XAI. Our framework can identify early signs of drift by combining it with multivariate drift detection and confidence based performance monitoring methods. 
    \item We show through systematic experiments that local XAI methods can exhibit a stable feature-importance even when predictive performance deteriorates, highlighting that they cannot be used alone to identify covariate drift.
    \item Our proposed framework jointly tracks estimated predictive performance and the stability of model explanations, enabling the detection of covariate drift at an early stage that may not be reflected in performance metrics alone.
\end{itemize}


\section{Related Work}
\label{sec:relatedwork}
The two popular and widely adopted approaches for XAI are Local Interpretable Model-agnostic Explanations (LIME)~\cite{ribeiro2016should} and SHapley Additive exPlanations (SHAP)~\cite{lundberg2017unified}. These methods explain individual predictions of any classifier in an interpretable and faithful manner, by learning an interpretable model (e.g., linear model) locally around each prediction. The authors in~\cite{pelosi2025explainability} presented a taxonomy based on the technical and methodological foundations of explainability and interpretability with respect to concept and data drift. 
The authors in another study~\cite{lee2023xaidrift} propose an approach to detect drift that does not rely on class labels. Instead, it calculates a drift score based on a statistical metric derived from SHAP values, and uses this score to assess whether the monitored data distribution has changed. In a related line of research, the authors in~\cite{duckworth2021using, wang2023harmful}, demonstrated how XAI can be applied to track the influence of input features on prediction and changes in the importance of features over time. These shifts can be used as an additional metric to identify data drift and highlight emerging health risks.

Furthermore, the authors in~\cite{mougan2025explanation} propose a framework to identify the explanation shift using SHAP, which is defined as a statistical comparison between how predictions are explained from training data and how predictions in new data are explained. The proposed method operates on explanation distributions, providing more sensitive and explainable changes in interactions between distribution shifts and learned models, referred to as the discriminatory model. Additionally, the authors in \cite{muller2024open} benchmarked drift detection techniques focusing on univariate detection methods. We extend this research direction by introducing FMMO to identify models vulnerable to performance deterioration under covariate drift.

\section{Background}
\label{sec:background}
Before presenting our methodology, we first describe its key components individually.

\textbf{ML Models:}
Random Forests (RF)~\cite{biau2016random} and gradient boosted trees such as eXtreme Gradient Boosting (XGB)~\cite{chen2015xgboost} are widely used non-linear predictive models~\cite{lundberg2020local}. They belong to different tree-based ensembling paradigms bagging and boosting, respectively, which aggregate several models to form a more powerful predictor~\cite{sutton2005classification}. 


\textbf{Feature Drift Monitoring:}
We employ the DC from NannyML (NML)~\cite{nannyml} to track statistical changes in feature distributions. DC is a multivariate drift detection technique that uses a data reconstruction strategy based on Principal Component Analysis (PCA) for dimensionality reduction to identify multivariate feature drift, enabling the detection of subtle structural changes that univariate drift detection methods miss~\cite{muller2024open,rabanser2019failing}.

\textbf{Model Utilization Monitoring:}
We utilize the NML implementation of the CBPE to monitor the utilization of the model~\cite{nannyml}. The key idea of CBPE is to leverage the confidence of the model's predictions. For classifiers, this confidence is provided by the predicted probability that an instance belongs to a particular class. These confidence scores are then used to estimate classification performance. If the monitored model produces well-calibrated probabilities and the sample size is sufficiently large, CBPE can reliably approximate performance. 

\textbf{XAI Feature Attribution Monitoring:}
Post-hoc model agnostic approaches are popular in practice, where the target is to design a separate simple algorithm that can approximate the decision making process of any ML model. Due to the instability issue of the LIME explanations~\cite{zafar2021deterministic}, we will not be able to know whether the change in the explanations is due to random perturbation of the samples or due to drift. Therefore, we will continue our experiments with SHAP. SHAP is inspired by game theory, where it calculates the contribution of each feature to the prediction. SHAP assigns each feature a score that represents its influence on the prediction, calculated by considering all possible subsets of features used in the model~\cite{lundberg2020local}. SHAP offer various methods to generate explanations; in our experiments, however, we specifically focus on TreeExplainer~\cite{lundberg2018consistent} (hereafter denoted as TreeSHAP) because it produces stable explanations, allowing us to attribute any changes in explanations due to drift, is fully compatible with our tree-based ML models, and supports both local and global explanations, providing the flexibility to perform various analysis.

Although TreeSHAP provides locally faithful feature attributions for individual predictions, its explanatory scope is inherently instance based and may not fully capture population-level dependencies. In particular, features whose contributions are modest in local contexts, but exert substantial influence when aggregated across the data distribution may not be consistently highlighted by local attributions. Motivated by these limitations of local attribution methods, we complement TreeSHAP with Global Surrogate Model (GSM) in our experiments to capture population-level feature dependence and interaction effects that are not apparent from locally aggregated explanations alone. The GSM is a model-agnostic interpretable model that approximates the predictions of a complex black-box ML model without requiring access to data and predictions of the black-box ML model~\cite{molnar2025}. 


\section{Methodology}
\label{sec:method}
\begin{figure*}[ht]
    \centering
    \includegraphics[scale=0.75]{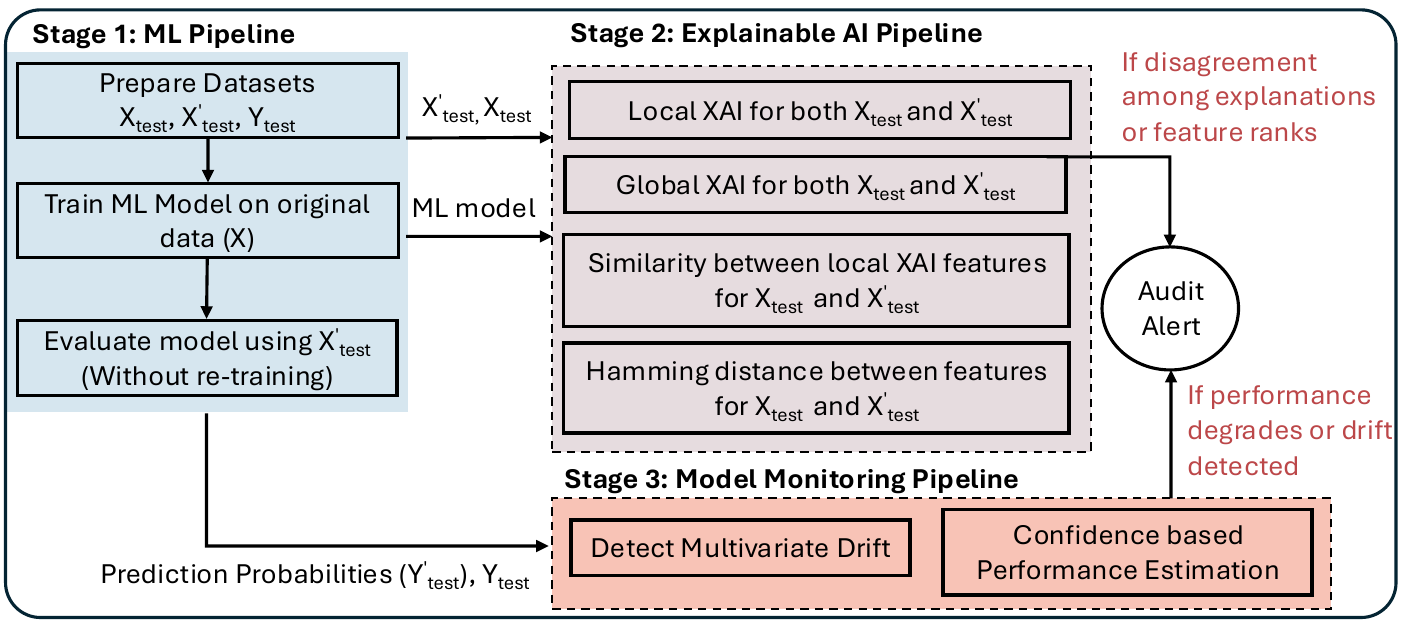}
    \caption{High-level block diagram of the proposed Framework for Model Monitoring and Observability (FMMO). It consists of three stages which includes 1) the ML pipeline, 2) the XAI pipeline, and 3) the ML model monitoring pipeline.}
    \label{fig:method}
\end{figure*}   
In this section, we present our Framework for Model Monitoring and Observability (FMMO). It consists of three stages as illustrated in Figure~\ref{fig:method}. The three stages of the FMMO are 1) the ML pipeline, 2) the XAI pipeline, and 3) the ML model monitoring pipeline. In stage one, we chose two tree-based methods (RF and XGB) due to a different approach to ensemble trees. Also, obtaining important global features from tree-based models is straightforward. In stage two, we choose TreeSHAP as the local XAI framework and Decision Tree (DT) as a GSM. We chose two XAI methods, as TreeSHAP approximates the model's behavior in the local vicinity, while a GSM approximates the black-box model's behavior with an interpretable decision tree to reveal its global decision logic. In stage three, we adopted two methods from NML to identify multivariate drift detection and CBPE. Our proposed framework jointly tracks estimated predictive performance and the stability of explanations, enabling the detection of covariate drift at an early stage with explanations that may not be reflected in performance metrics alone, and helps to decide whether the ML model requires an audit. The Algorithm~\ref{algo:method} formally outlines the proposed method. It takes as input the dataset ($D$), the ML model ($M$), the XAI tool ($E$), the model monitoring framework ($MM$), the type of noise, and $k$ (used to select the top features). In addition, $\beta$ and $\zeta$ are user-specified threshold parameters that determine when alerts are triggered. These thresholds govern how stringent the alert conditions are. Choosing $\beta = 1$ and $\zeta = 1$ enforces exact agreement between feature vectors and rankings, which can be too restrictive in practical deployment scenarios. Based on these inputs, it outputs a signal, along with explanations, indicating whether an audit of the ML model is necessary. 

\begin{algorithm}[tb]
\caption{A Framework for Model Monitoring and Observability (FMMO)}
\label{algo:method}
\textbf{Input}: Dataset $D = \{\bf X, \bf y\}$, trained model $M$, explainers $E$, model monitoring method $MM$, type\_of\_noise, k, $\beta$, $\zeta$\\
\textbf{Output}: Decision with explanation

\begin{algorithmic}[1] 
\STATE Split $D$ into reference set $D_{ref}$ and validation set $D_{val}$.
\STATE Train model $M$ on $D_{ref}$ 
\STATE Compute baseline metrics (Accuracy, ROC AUC) on $D_{val}$
\IF {$type\_of\_noise = \text{Gaussian}$}
\STATE Add Gaussian noise using Equation~\ref{eq:gaussian} in $D_{val}$ to produce $D'_{val}$.
\ELSE
\STATE Add Uniform noise using Equation~\ref{eq:uniform} in $D_{val}$ to produce $D'_{val}$.
\ENDIF
\STATE Evaluate $M$ on $D'_{val}$ without retraining
\STATE Record performance metrics (Accuracy, ROC AUC) using $D'_{val}$
\STATE $E_{b} =$ Generate explanations using $E$ for $D_{val}$
\STATE $E_{d} =$ Generate explanations using $E$ for $D'_{val}$
\STATE $\mathcal{J} =$ Compute Jaccard similarity $\mathcal{J} (E_b, E_d)$ using Equation~\ref{eq:jaccard} among top $k$ explanations
\STATE $\alpha =$ Compute feature rank agreement $\alpha (E_{baseline}, E_{drift})$ using Equation \ref{eq:hamming} among top $k$ explanations
\STATE $d$ = Detect multivariate drift for $D'_{val}$ to identify distributional changes using $MM$ and capture alert 
\STATE $p$ = Model utilization monitoring for $D'_{val}$ using $MM$ and capture alert
\IF {$\mathcal{J} < \beta$ \textbf{or} $\alpha < \zeta$ \textbf{or} $d$ is $True$ \textbf{or} $p$ is $True$}
\STATE \textbf{return} ML model audit required along with explanation
\ELSE
\STATE \textbf{return} ML model audit not required along with explanation
\ENDIF

\end{algorithmic}
\end{algorithm}

\section{Experiment Design and Results}
\label{sec:results}
This section presents the experiment design, reports FMMO results, and provides a detailed analysis. In our experiments, we injected covariate drift into the datasets and trained the RF and XGB classifiers on the original data ($X$) and tested it on the drifted data ($X'$). We then used TreeSHAP to generate local explanations and computed the importance of features for RF and XGB on both $X$ and $X'$. To assess whether the set of important features changed, we compared the top $K$ features using the Jaccard similarity and their ranks using the Hamming distance, as defined in equations~\ref{eq:jaccard} and \ref{eq:hamming}, respectively. We performed the same procedure with the global XAI method (GSM) to obtain feature importance. Lastly, we used NML's DC to verify that the drift injected in $X'$ was strong enough to be reliably detected.

\textbf{Dataset and Pre-processing:} The experiments used a pre-processed version of the real-world Consumer Expenditure Survey (CES)~\cite{bls2020ces}, benchmark datasets including  Correctional Offender Management Profiling for Alternative Sanctions (COMPAS)~\cite{compas}, German Credit (GC)~\cite{blake1999repository}, and Communities \& Crime (CC)~\cite{redmond2002data}, and a synthetic data set generated using scikit-learn~\cite{kramer2016scikit}. A summary of these datasets is presented in Table~\ref{tab:data}. Each dataset was pre-processed by removing columns that are irrelevant or serve only as identifiers, applying one-hot encoding to categorical variables while preserving their categorical type for drift evaluation, normalizing the numerical features, and handling missing values. CES-NA uses the same underlying data as CES, but is divided by years 2017–2019 for training and 2020 for testing. Because it naturally reflects the impact of COVID-19, we did not inject additional noise into CES-NA for our experiments, unlike CES, where we mixed this dataset.


\textbf{Data Drift Generation:} In our experiments, we used covariate drift by adding noise from Gaussian and Uniform distributions as defined in equations~\ref{eq:gaussian} and \ref{eq:uniform} to synthetically induce drift in the data. The covariate drift describes a situation in which the distribution of independent variables $P(X)$ differs between the training and the test dataset, while the dependent variable $P(Y)$ remains the same~\cite{shimodaira2000improving, mallick2022matchmaker}. The consideration of synthetic drift enables us to precisely control the drift intensity to measure the sensitivity threshold for TreeSHAP and GSM DT, which is more difficult to isolate in wild data. 

\begin{equation}
\mathbf{X}' = \mathbf{X} + \boldsymbol{\varepsilon},
\quad
\boldsymbol{\varepsilon} \sim \mathcal{N}(\boldsymbol{\mu}, \boldsymbol{\Sigma})
\label{eq:gaussian}
\end{equation}

where $\boldsymbol{\mu}$ denotes the mean vector and $\boldsymbol{\Sigma}$
denotes the Gaussian noise covariance matrix. In our experiments,
we used $\boldsymbol{\mu}=\mathbf{0}$ and $\boldsymbol{\Sigma}=\mathbf{I}$.

\begin{equation}
\mathbf{X}' = \mathbf{X} \cdot U + s \cdot d,
\quad
U \sim \mathcal{U}(a,b),
\quad
d \in \{-1,+1\}
\label{eq:uniform}
\end{equation}
where $a$ is the lower bound and $b$ is the upper bound. We used $a = 1.5$ and $b = 3.5$ to simulate a strong asymmetric covariate drift by independently rescaling each feature using random factors sampled from a Uniform distribution $\mathcal{U}(1.5, 3.5)$ and apply directional mean shifts parameterized by a Rademacher variable $d \in \{-1, +1\}$, where $s$ controls the drift intensity. The parameter $s$ is calibrated so that the performance of the model is substantially reduced. These parameter choices are specifically designed to enforce a strong covariate drift in the datasets. The effect of injected drift is illustrated in Figure~\ref{fig:drift_example}. The green line shows the reference mean, and the light green band marks the reference mean $\pm 1$  standard deviation as the stability region. The red line represents the current mean age per bin, with the red shaded area indicating its spread. In the figure at the bottom, the red line (current mean) is not approaching the green line (reference mean) and does not stabilize within the reference central region, indicating a sustained shift in the underlying data distribution.
After introducing drift into all datasets, both ML models were evaluated without retraining to examine the impact on model performance and explanations of the XAI methods.

\textbf{Model Training:} The RF and XGB classifiers were selected as predictive models due to their inherent robustness and compatibility with interpretability, which are critical attributes in the realm of predictive modeling. First, both classifiers were trained on the original data using a 70:30 ratio for training and testing, respectively. We evaluated ML classifiers using key performance metrics such as Receiver Operating Characteristic Area Under the Curve (ROC AUC)~\cite{hanley1982meaning}. The ROC AUC is a commonly used measure to assess the effectiveness of classification models. It is especially valuable for imbalanced datasets and scenarios requiring several decision thresholds. In our results, we refer to it as the baseline ROC AUC.

\textbf{Evaluation Metrics:} To obtain explanations, we used post-hoc interpretability methods, specifically TreeSHAP and DT as GSM. For TreeSHAP, we computed mean absolute values for all features in both $\textbf{X}_{test}$ and $\textbf{X}'_{test}$ to assess the stability of the importance of the feature. The importance of the features of the three methods in both $\textbf{X}_{test}$ and $\textbf{X}'_{test}$ was assessed by comparing the overlap of features using Jaccard similarity ($\mathcal{J}$) and rank agreement of features ($\alpha$) using the normalized Hamming distance ($d_\mathrm{H}$), as defined in Equations~\ref{eq:jaccard} and \ref{eq:hamming}, respectively.

\begin{equation}
\mathcal{J}(\bf a, \bf b) = \frac{|\bf a \cap \bf b|}{|\bf a \cup \bf b|}
\label{eq:jaccard}
\end{equation}

where $\bf a$ and $\bf b$ are ranked feature vectors.


\begin{equation}
\alpha = 1 - d_{\mathrm{H}}(\mathbf{a}, \mathbf{b}),
\quad
d_\mathrm{H}(\bf a, \bf b) = \frac{1}{k}\sum_{i=1}^{k}
\mathds{1}\!\left[a_i \neq b_i\right] 
\label{eq:hamming}
\end{equation}
where $\bf a$ and $\bf b$ are ranked feature vectors, $k$ is the length of the vectors, and $\mathds{1}[\cdot]$ denotes the indicator function, which equals $1$ if the condition holds and $0$ otherwise.



Both $\alpha$ and $\mathcal{J}$ are similarity metrics, but capture different notions of similarity. The $\alpha$ a position-wise distance measure that counts how many positions differ between two objects of equal length, whereas $\mathcal{J}$ is a set-based similarity measure that quantifies the overlap between two sets with respect to their union. Together, they offer complementary perspectives, one on consistency of ranks and the other on shared membership. We additionally evaluated group-wise FPR and Accuracy, using Equations~\ref{eq:fpr} and \ref{eq:accuracy}, to quantify drift effects on the protected attributes.
\begin{equation}
\text{FPR}_g = \frac{FP_g}{FP_g + TN_g}
\label{eq:fpr}
\end{equation}

\begin{equation}
\text{Acc}_g = \frac{TP_g + TN_g}{N_g}
\label{eq:accuracy}
\end{equation}

where $TP_g$, $FP_g$, $TN_g$, and $FN_g$ denote the number of true positives, false positives, true negatives, and false negatives for group $g$, and $N_g$ is the total number of samples in group $g$.

\subsection{Results}
Before discussing the results in detail, we briefly describe the content and interpretation of Table~\ref{tab:baseline}, Table~\ref{tab:results}, Figure~\ref{fig:nml_ces} and Figure~\ref{fig:nml_compas}. Both Figures~\ref{fig:nml_ces} and \ref{fig:nml_compas} present the same information for different experiments on two different datasets.

\begin{itemize}
    \item Table~\ref{tab:baseline} reports the accuracy and ROC AUC of the ML models evaluated on test data. A baseline result provides a reference for evaluating the ML model’s performance on drifted data.
    
    \item Table~\ref{tab:results} reports the noise added for covariate drift, the ML model used, ROC AUC, the Jaccard similarity ($\mathcal{J}$) and the feature rank agreement ($\alpha$) between the original test set $X_{\text{test}}$ and the test set with drift $X'_{\text{test}}$ computed for features obtained using different top-$k$ settings from TreeSHAP and Decision Tree (DT) for top-$5$ features. In the ROC AUC columns, bold values indicate a slight performance decrease of the ML model, while bold red values denote a substantial drop. In the other columns, $\mathcal{J}$ quantifies explanation stability and $\alpha$ quantifies feature rank stability. Values of $\mathcal{J}$ and $\alpha$ close to $0$ indicate high dissimilarity between explanations and strong disagreement in feature rankings, whereas a value of $100$ denotes perfect agreement in feature rankings and highly stable, consistent explanations. The alert columns indicate whether the FMMO triggered an alert for the audit of the ML model.

    \item Figures~\ref{fig:nml_ces} and \ref{fig:nml_compas} present the FMMO results for the CES and COMPAS dataset under added Gaussian noise, comprising six panels. The first panel~\ref{fig:nml_dcc_ces} shows the DC's drift detection report, the second~\ref{fig:nml_dcc_ces} shows the CBPE report, the third~\ref{fig:dt_baseline_ces} and fourth~\ref{fig:dt_drift_ces} display GSM-based explanations, and the fifth~\ref{fig:dt_shap_baseline_ces} and sixth~\ref{fig:dt_shap_drift_ces} present TreeSHAP-based explanations.
    
    \item In Figure~\ref{fig:nml_dcc_ces}, the x-axis indexes sequential data chunks, each representing a batch of observations processed over time. The y-axis shows the ROC AUC of a domain classifier trained to distinguish reference from analysis data, capturing changes in the joint feature distribution rather than changes in individual features. The Reference region on the left indicates the baseline behavior and the Analysis region on the right corresponds to the analysis data. Red markers indicate drift alerts, triggered when the ROC AUC exceeds or drops the preset threshold shown by the red dashed lines.

    \item Figure~\ref{fig:nml_cbpe_ces} shows model's performance monitored using CBPE across sequential data chunks represented on the x-axis. The y-axis shows the estimated ROC AUC of the model. The shaded region on the left corresponds to the reference period used to calibrate the estimator, while the right region represents the analysis period. The red dashed lines serve as a threshold for acceptable performance limits. The blue dashed line shows the estimated model performance for each chunk. Crossing these lines would indicate a potential performance issue.

    \item Figures~\ref{fig:dt_baseline_ces} and \ref{fig:dt_drift_ces} present the top $5$ features of DT before and after covariate drift injection. The x-axis indicates the features, the y-axis shows their importance, the sky blue bars correspond to the baseline, and the salmon pink bars to the drifted data.

    \item Figures~\ref{fig:dt_shap_baseline_ces} and \ref{fig:dt_shap_drift_ces} present the top $5$ features of TreeSHAP before and after drift injection. The x-axis indicates the features, the y-axis shows their aggregated TreeSHAP values, the blue bars correspond to the baseline, and the brown bars to the drifted data.
\end{itemize}

\begin{table}[t]
\centering
  \small
    \caption{Baseline performance metrics on the test dataset}
    \begin{tabular}{lccc}
        \toprule
        \bf Dataset & \bf ML Model & \bf Accuracy & \bf ROC AUC \\
        \midrule

        \multirow{2}{*}{CES – NA}&RF&$89.85$&$85.90$ \\
        \cmidrule{2-4}
        &XGB&$90$&$85.55$ \\
        
        \midrule
        
        \multirow{2}{*}{CES}&RF&$91.30$&$87.00$ \\
        \cmidrule{2-4}
        &XGB&$90.88$&$85.43$ \\
        
        \midrule
        
        \multirow{2}{*}{COMPAS}&RF&$65.60$&$70.80$ \\
        \cmidrule{2-4} 
        &XGB&$67.77$&$73.93$ \\ 
        
         \midrule
        \multirow{2}{*}{CC}&RF&$82.30$&$90.16$ \\
        \cmidrule{2-4}
        &XGB&$81.30$&$89.35$ \\

        \midrule
        
        \multirow{2}{*}{GC}&RF&$72.00$&$75.70$ \\
        \cmidrule{2-4}
        &XGB&$73.66$&$71.98$ \\

        \midrule
        
        \multirow{2}{*}{Synthetic}&RF&$89.54$&$94.71$ \\
        \cmidrule{2-4}
        &XGB&$91.00$&$95.95$ \\

        \bottomrule        
    \end{tabular}
    \label{tab:baseline}
\end{table}

\begin{table}[t]
\centering
  \small
    \caption{Summary of Datasets}
    \label{tab:data}
    \begin{tabular}{lccl}
        \toprule
        \bf Dataset & \bf Observations & \bf Features & \bf Type \\
        \midrule  
        COMPAS & $7,214$ &  $53$ &   Binary \\ 
        \midrule
        GC & $1,000$ & $30$ & Binary \\
        \midrule
        CC & $2,215$ & $128$ & Binary \\ 
        \midrule
        CES & $16,789$ &  $26$ &   Binary \\ 
        \midrule
        Synthetic & $25,000$ & $20$ & Binary \\ 
        \bottomrule
     \end{tabular}

\end{table}

\begin{table}[t]
\centering
  \small
    \caption{Performance Metrics for COMPAS Dataset using Random Forest (RF)}
    \label{tab:compas_rf_group_metrics}
    \begin{tabular}{l l l c c c}
    \toprule
    \textbf{Dataset} & \textbf{Drift} & \textbf{Group} & \textbf{Settings} & \textbf{FPR} & \textbf{Accuracy} \\
    \midrule
    
    \multirow{8}{*}{COMPAS} & \multirow{4}{0.2cm}{High} 
    & \multirow{2}{*}{Protected} 
    & Baseline & $37.84$ & $63.93$ \\
    &&  & Drift & $ \textbf{70.39}$ & $ \textbf{59.78}$ \\
    \cmidrule(lr){3-6}
    && \multirow{2}{*}{Privileged} 
    & Baseline & $10.17$ & $67.98$ \\
    &&  & Drift & $ \textbf{36.17}$ & $63.63$ \\

     \cmidrule(lr){2-6}
     
    & \multirow{4}{0.2cm}{Low} 
    & \multirow{2}{*}{Protected} 
    & Baseline & $37.84$ & $63.93$ \\
    &&  & Drift & $ \textbf{44.50}$ & $ \textbf{63.06}$ \\
    \cmidrule(lr){3-6}
    && \multirow{2}{*}{Privileged} 
    & Baseline & $10.17$ & $67.98$ \\
    &&  & Drift & $ \textbf{13.23}$ & $67.70$ \\
    
    \bottomrule
    \end{tabular}

\end{table}

\subsubsection{\textbf{Baseline Performance}}
RF and XGB were trained on the original datasets discussed in Table~\ref{tab:data} using a data split ratio $70:30$ for training ($\textbf{X}_{train}, \textbf{y}_{train}$) and testing ($\textbf{X}_{test}, \textbf{y}_{test}$), respectively. We evaluated ML models by computing Accuracy and ROC AUC and refer to it as baseline performance and results are summarized in Table~\ref{tab:baseline}. Both XGB and RF score more than $70$\% accuracy on all datasets except the COMPAS dataset, where RF scored $65.6$\% accuracy and XGB scored $67.77$\% accuracy. Their performance could be further improved by hyperparameter tuning, which we leave for future work since our focus here is on evaluating whether XAI methods can assist in identifying covariate drift. 


        
        
        
        
        

        

        


\subsubsection{\textbf{Local XAI}}
\begin{table*}[ht]
\centering
\small
\caption{Comparative summary of ROC AUC, explanation stability and features rank agreement under covariate drift.}
\begin{tabular}{l l c c *{6}{c} *{2}{c} c}
\toprule

\bf Dataset & \bf Noise & \bf ML Model & \bf ROC AUC
& \multicolumn{6}{c}{\bf{TreeSHAP}}
& \multicolumn{2}{c}{\bf{DT}}

& \bf {Alert} \\

\cmidrule(lr){5-10}
\cmidrule(lr){11-12}

& & & 
& \multicolumn{2}{c}{$k=3$}
& \multicolumn{2}{c}{$k=5$}
& \multicolumn{2}{c}{$k=10$}
& \multicolumn{2}{c}{$k=5$}
& \\
\cmidrule(lr){5-6}
\cmidrule(lr){7-8}
\cmidrule(lr){9-10}
\cmidrule(lr){11-12}

& & & 
& $\mathcal{J}$ & $\alpha$
& $\mathcal{J}$ & $\alpha$
& $\mathcal{J}$ & $\alpha$
& $\mathcal{J}$ & $\alpha$
& \\

\midrule

\multirow{2}{*}{CES - NA} & \multirow{2}{4em}{None} & RF  & $\textbf{84.54}$
 & $100$ & $100$
 & $100$ & $100$
 & $100$ & $ \textbf{80}$

 & $\textbf{66.67}$ & $ \textbf{20}$    

 &  TRUE \\

 &  & XGB &  $\textbf{82.90}$
 & $100$ & $\textbf{33.33}$
 & $\textbf{66.67}$ & $\textbf{40}$
 & $100$ & $\textbf{20}$

 & $ \textbf{66.67}$ & $\textbf{60}$   

 &  TRUE \\

 \midrule
 
\multirow{4}{*}{CES} & \multirow{2}{4em}{Gaussian} & RF  & $86.99$
 & $100$ & $100$
 & $100$ & $100$
 & $100$ & $\textbf{80}$

 & $ \textbf{25}$ & $\textbf{20}$    

 &  TRUE \\

 &  & XGB &  $85.28$
 & $100$ & $100$
 & $100$ & $100$
 & $100$ & $100$

 & $\textbf{66.67}$ & $\textbf{40}$   

 &  TRUE \\

 \cmidrule {2-13}

 & \multirow{2}{4em}{Uniform} & RF  & $ 86.55$
 & $100$ & $\textbf{33}$ 
 & $100$ & $\textbf{33}$ 
 & $100$ & $\textbf{20}$ 

 & $\textbf{42.86}$ & $\textbf{0}$ 
 
 &  TRUE \\

 &  & XGB & $\textbf{79.47}$
 & $100$ & $100$
 & $100$ & $100$
 & $\textbf{66.67}$ & $\textbf{50}$

 & $100$ & $\textbf{20}$
 
 &  TRUE \\

 \midrule

\multirow{4}{*}{COMPAS} & \multirow{2}{4em}{Gaussian} & RF  & $\textbf{70.55}$
 & $100$ & $100$
 & $100$ & $100$
 & $100$ & $100$

 & $\textbf{66.67}$ & $\textbf{60}$   
 
 &  TRUE \\

 &  & XGB & $\textbf{72.66}$
 & $100$ & $100$
 & $100$ & $100$
 & $100$ & $100$

 & $100$ & $\textbf{60}$  

 &  TRUE \\

 \cmidrule {2-13}

 & \multirow{2}{4em}{Uniform} & RF  & $\textbf{70.66}$
 & $100$  & $\textbf{33.33}$
 & $100$ & $\textbf{60}$
 & $100$ & $\textbf{40}$

 & $\textbf{42.86}$ & $\textbf{60}$    
 
 &  TRUE \\

 &  & XGB & $\textbf{70.99}$
 & $100$ & $100$
 & $100$ & $\textbf{60}$
 & $\textbf{81.82}$ & $\textbf{70}$

 & $\textbf{66.67}$ & $\textbf{40}$ 
 
 &  TRUE \\

 \midrule

\multirow{4}{*}{CC} & \multirow{2}{4em}{Gaussian} & RF  & $\textbf{89.20}$
 & $100$ & $\textbf{33.33}$
 & $100$ & $\textbf{60}$
 & $100$ & $\textbf{60}$

 & $\textbf{42.86}$ & $\textbf{20}$  

 &  TRUE \\

 &  & XGB & $\textbf{89.16}$
 & $100$ & $100$
 & $100$ & $\textbf{80}$
 & $100$ & $\textbf{80}$

 & $\textbf{66.67}$ & $\textbf{80}$ 
 
 &  TRUE \\

 \cmidrule {2-13}

 & \multirow{2}{4em}{Uniform} & RF  & $\textbf{81.09}$
 & $ \textbf{20}$ & $\textbf{0}$
 & $\textbf{55.56}$ & $\textbf{20}$
 & $\textbf{66.67}$ & $\textbf{10}$

 & $\textbf{11.11}$ & $\textbf{0}$ 
 
 &  TRUE \\

 &  & XGB & $\textbf{81.40}$
 & $\textbf{20}$ & $\textbf{0}$
 & $\textbf{66.67}$ & $\textbf{20}$
 & $\textbf{42.86}$ & $\textbf{10}$

 & $\textbf{25}$ & $\textbf{0}$    

 &  TRUE \\
 
 \midrule

\multirow{4}{*}{GC} & \multirow{2}{4em}{Gaussian} & RF  & $75.70$
 & $100$ & $100$
 & $100$ & $100$
 & $100$ & $\textbf{80}$

 & $100$ & $\textbf{60}$  

 &  TRUE \\

 &  & XGB & $\textbf{71.13}$
 & $100$ & $100$
 & $100$ & $100$
 & $100$ & $\textbf{80}$

 & $\textbf{66.67}$ & $\textbf{0}$
 
 &  TRUE \\

 \cmidrule {2-13}

 & \multirow{2}{4em}{Uniform} & RF  & $\textbf{72.98}$
 & $\textbf{50}$ & $\textbf{66.67}$
 & $100$ & $\textbf{60}$
 & $\textbf{66.67}$ & $\textbf{40}$

 & $\textbf{42.86}$ & $\textbf{40}$   

 &  TRUE \\

 &  & XGB & $\textbf{67.93}$
 & $100$ & $\textbf{0}$
 & $\textbf{66.67}$ & $\textbf{20}$
 & $\textbf{81.82}$ & $\textbf{10}$

 & $\textbf{42.86}$ & $\textbf{60}$ 

 &  TRUE \\
  
 \midrule

\multirow{4}{*}{Synthetic} & \multirow{2}{4em}{Gaussian} & RF  & $\textbf{84.55}$
 & $100$ & $100$
 & $100$ & $100$
 & $\textbf{81.82}$ & $\textbf{90}$

 & $\textbf{66.67}$ & $\textbf{80}$

 &  TRUE \\

 &  & XGB & $\textbf{83.97}$
 & $100$ & $100$
 & $\textbf{66.67}$ & $\textbf{80}$
 & $\textbf{81.82}$ & $\textbf{40}$

 & $\textbf{66.67}$ & $\textbf{40}$  

 &  TRUE \\

 \cmidrule {2-13}

 & \multirow{2}{4em}{Uniform} & RF  & $\textbf{90.79}$
 & $100$ & $100$
 & $\textbf{66.67}$ & $\textbf{80}$
 & $\textbf{42.86}$ & $\textbf{40}$

 & $\textbf{0}$ & $\textbf{0}$  

 &  TRUE \\

 &  & XGB & $\textbf{92.58}$
 & $\textbf{50}$ & $\textbf{66.67}$
 & $\textbf{25}$ & $\textbf{40}$
 & $\textbf{42.86}$ & $\textbf{20}$

 & $\textbf{66.67}$ & $\textbf{80}$ 

 &  TRUE \\
 
\bottomrule
\end{tabular}
\label{tab:results}
\end{table*}

\begin{figure*}[ht]
    \centering
    \begin{subfigure}{0.48\linewidth}
        \centering
        \includegraphics[scale=0.35]{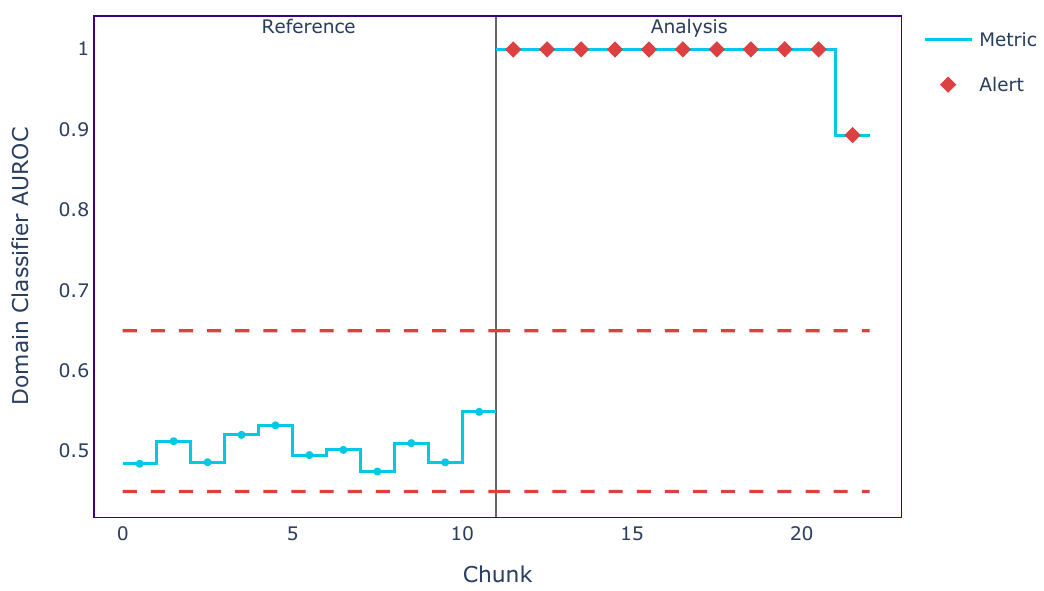}
        \caption{Multivariate drift detection (DC) }
        \label{fig:nml_dcc_ces}
    \end{subfigure}
    \hspace{1em}
    \begin{subfigure}{0.48\linewidth}
        \centering
        \includegraphics[scale=0.35]{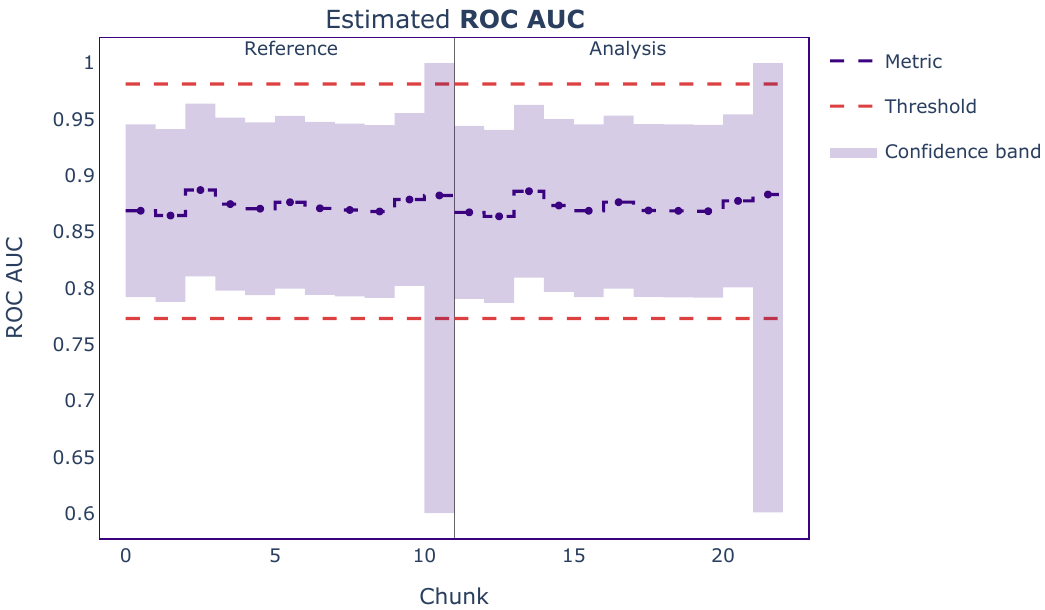}
    \caption{Estimated performance (CBPE)}
    \label{fig:nml_cbpe_ces}
    \end{subfigure}
    
    \begin{subfigure}{0.48\linewidth}
        \centering
        \includegraphics[scale=0.4]{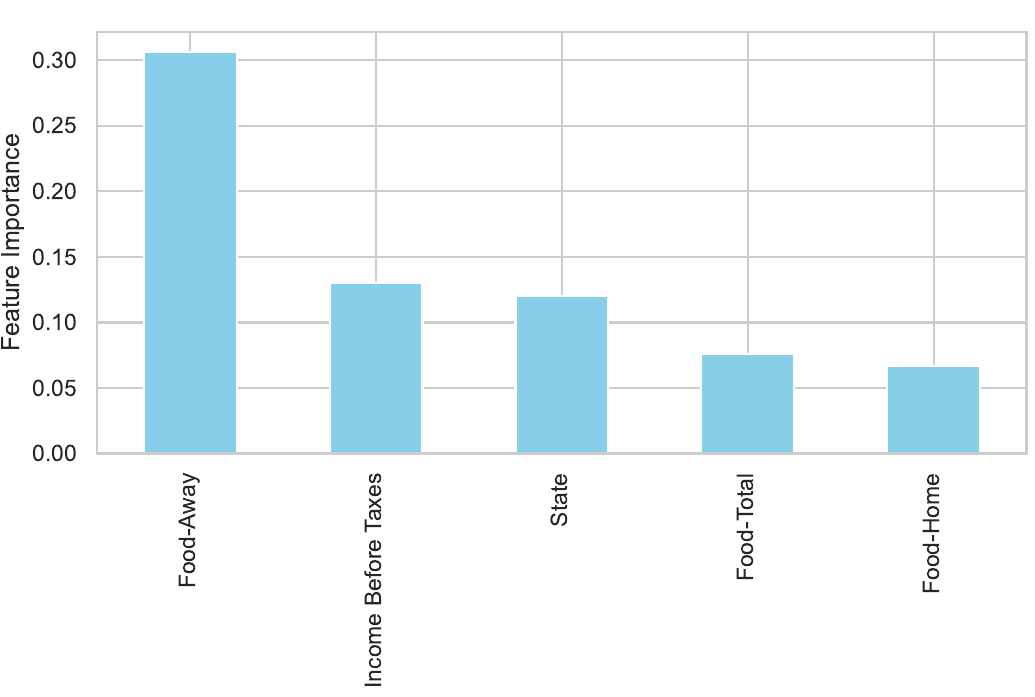}
        \caption{DT - Top 5 features before drift) }
        \label{fig:dt_baseline_ces}
    \end{subfigure}
    \hspace{1em}
    \begin{subfigure}{0.48\linewidth}
        \centering
        \includegraphics[scale=0.4]{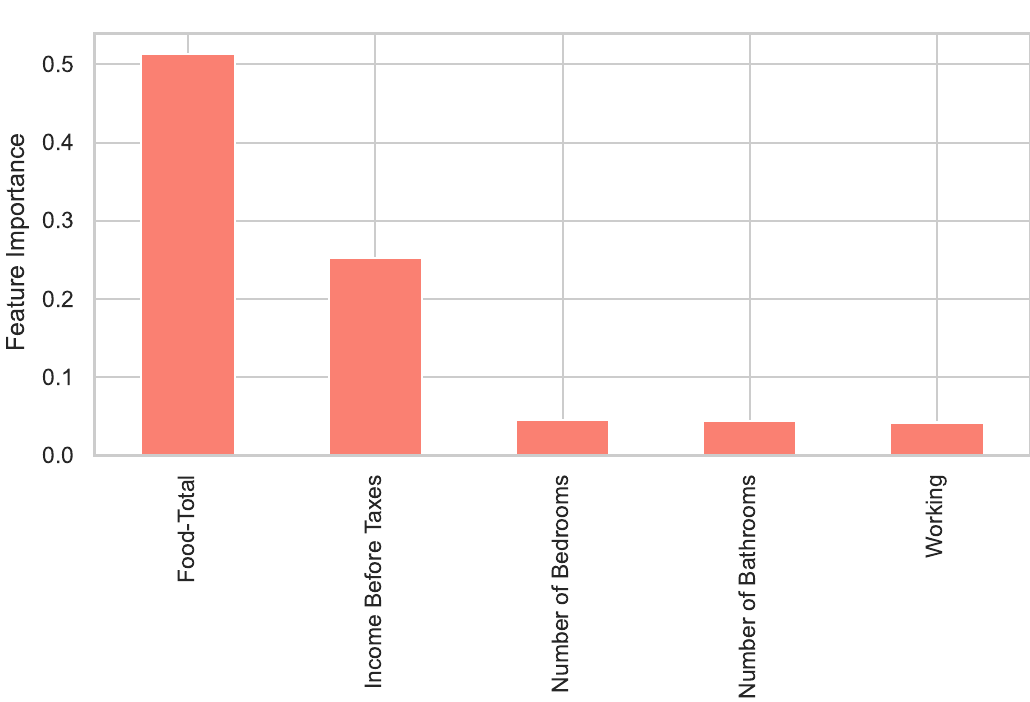}
    \caption{DT - Top 5 features after drift}
    \label{fig:dt_drift_ces}
    \end{subfigure}

    \begin{subfigure}{0.48\linewidth}
        \centering
        \includegraphics[scale=0.4]{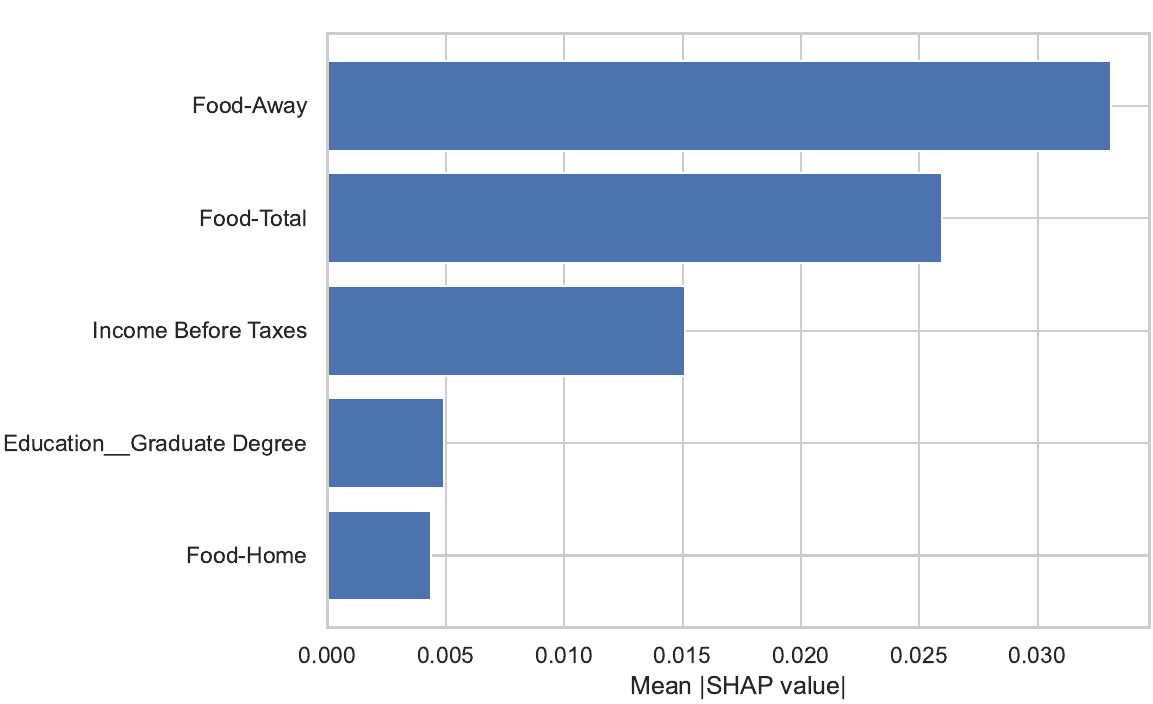}
        \caption{TreeSHAP - Top 5 features before drift }
        \label{fig:dt_shap_baseline_ces}
    \end{subfigure}
    \hspace{1em}
    \begin{subfigure}{0.48\linewidth}
        \centering
        \includegraphics[scale=0.4]{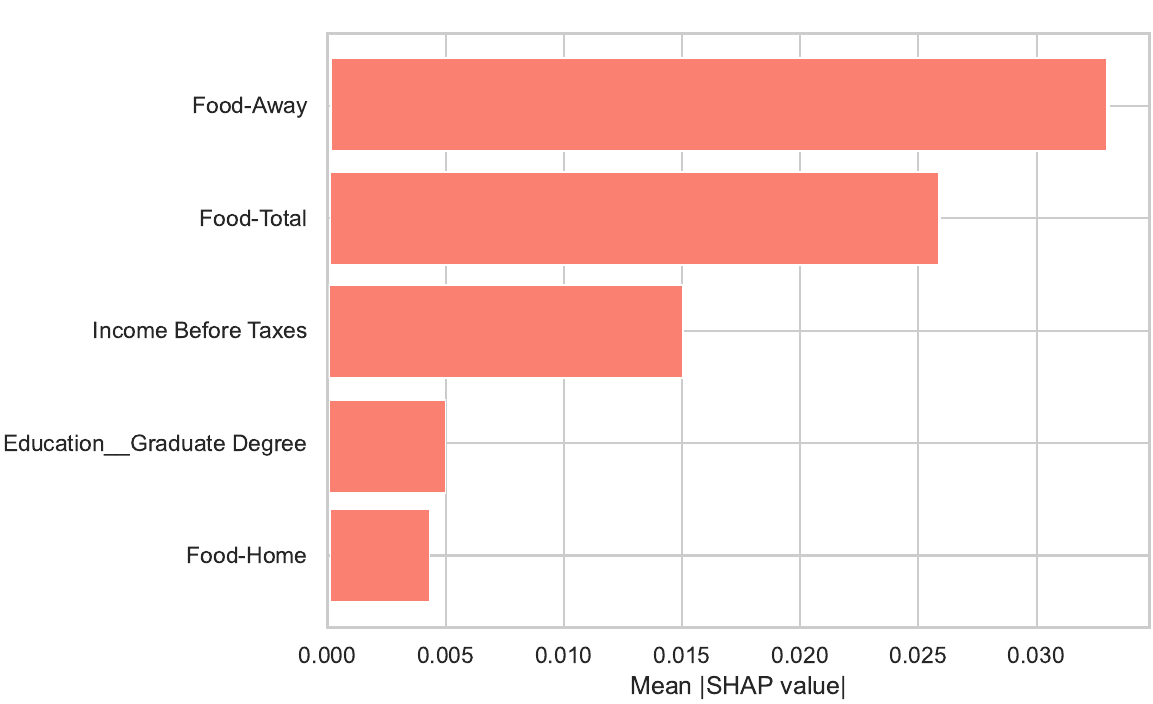}
    \caption{TreeSHAP - Top 5 features after drift}
    \label{fig:dt_shap_drift_ces}
    \end{subfigure}
    
    \caption{FMMO reports for CES dataset when Gaussian noise was added}
    \label{fig:nml_ces}
\end{figure*}

In our experiments, after splitting the dataset into training and test sets, we added noise to all test samples to synthetically induce covariate drift, denoting the resulting test set as $\textbf{X}'_{test}$, which is used to evaluate ML models without retraining. The results of Table~\ref{tab:results} and Figure~\ref{fig:nml_ces} show that when Gaussian noise was added to the numerical features of the CES dataset, the ROC AUC remained unchanged for both RF and XGB, $86.99$\% and $85.28$\%, respectively. The values of $\mathcal{J}$ and $\alpha$ for the top $3$, $5$ and $10$ also remained at $100$\% except $\alpha=80$\% for RF when $k=10$, indicating that TreeSHAP was not affected by covariate drift. The top $5$ features extracted from RF using $\textbf{X}_{test}$ and drifted $\textbf{X}'_{test}$ are shown in Figures~\ref{fig:dt_shap_baseline_ces} and \ref{fig:dt_shap_drift_ces}, respectively. It can be observed that not only the top five features remained unchanged, but their average TreeSHAP values also remained the same, indicating that TreeSHAP was not affected by drift. It is possible that aggregated TreeSHAP attributions remain insensitive to distributional changes that do not affect the predictive behavior of the ML model as it approximates the model behavior in the local vicinity by perturbing data around an instance $\textbf{x} \in \textbf{X}_{test}$. 

As an additional analysis, we computed the CBPE and multivariate drift detection using DC. 
As we can see in Figure~\ref{fig:nml_cbpe_ces}, the CBPE also remained stable, as the ROC AUC remains in the confidence band and did not exceed the threshold, suggesting that there was no degradation in the performance of the ML model. In contrast, DC detected drift successfully as shown in Figure~\ref{fig:nml_dcc_ces}. For each data chunk, the ROC AUC exceeded the threshold established from the reference data and DC triggered an alert. 

We repeated the same experiment by adding uniform noise. When uniform noise was added to the same features of the CES dataset, the ROC AUC of the RF decreased from $87$\% to $86.55$\%, which is not a significant decrease, and $\mathcal{J}$ remained $100\%$ for the top $3$, $5$ and $10$, suggesting that the model continued to rely on the same dominant features. However, the feature rankings changed significantly with $\alpha = 33$\% for the top $3$ and $5$ features and $\alpha = 20$\% for the top $10$ features, suggesting that TreeSHAP reacted slightly to distributional changes. The CBPE also remained stable, as the ROC AUC remains in the confidence band and did not exceed the threshold, suggesting that there was no degradation in the performance of the ML model. However, DC detected drift successfully.

Similarly, when Gaussian noise was added to the CC dataset, the ROC AUC of the RF dropped insignificantly and $\mathcal{J}$ remained $100\%$. However, the rank of important features differed significantly from $\alpha=33.33\%$ for the top $3$ and $\alpha=60\%$ for the $5$ and $10$ features, indicating that TreeSHAP was affected by drift. When the same experiment was repeated by adding uniform noise, the ROC AUC of the RF decreased significantly from $90.16$\% to $81.09$\% so $\mathcal{J}$. Additionally, both DC and CBPE raised alerts that indicated not only decreased performance but also detected drift. 

Furthermore, when we added Gaussian noise to other datasets such as COMPAS and GC, the ROC AUC decreased slightly, while the TreeSHAP feature rankings and their overlap remained unchanged with $\mathcal{J} = 100$\% for both both RF and XGB classifiers, as shown in Table~\ref{tab:results}. The only exception was $\alpha$, which was slightly affected when $k=10$, although $80$\% of the features still retained their rank. In contrast, the addition of uniform noise caused substantial changes in TreeSHAP rankings and a marked drop in $\mathcal{J}$, except for the COMPAS dataset, which showed mixed results.

These scenarios indicate that TreeSHAP may not react to certain feature-level distributional drifts and may not be used alone to identify covariate drifts. This is not the failure of the model, but a fundamental limitation of TreeSHAP as it explains model behavior locally rather than changes in the underlying data distribution. This limitation motivated us to explore global XAI models to identify covariate drift.



\subsubsection{\textbf{Global XAI}}
In our experiments, we used DT as a GSM to approximate the decision logic of black-box ML models. Repeating the experiments with DT revealed a strong sensitivity to both noise types. With Gaussian noise in CES dataset and RF as the base ML model, we obtained $\mathcal{J} = 25\%$ and $\alpha = 20\%$ for the top $5$ features using $\textbf{X}_{test}$ and $\textbf{X}'_{test}$, indicating that only $25$\% of the features overlapped and only $20$\% preserved their original rank. Under uniform noise, DT was significantly affected, with $\mathcal{J} = 42.86\%$ and $\alpha = 0\%$, meaning $42.86$\% of the features overlapped and no feature maintained their rank, as shown in Figure~\ref{fig:nml_ces}. Figure~\ref{fig:dt_baseline_ces} shows the top $5$ features extracted from DT using $\textbf{X}_{test}$ and Figure~\ref{fig:dt_drift_ces} shows the top $5$ features extracted from DT after training it on $\textbf{X}'_{test}$. This shows that DT is sensitive to shifts in data distributions that local XAI methods, specifically TreeSHAP, did not capture. Furthermore, using the CES-NA dataset and RF as the base ML model, we achieved $\mathcal{J} = 66.67\%$ and $\alpha = 20\%$ for the top $5$ features on $\textbf{X}_{test}$ and $\textbf{X}'_{test}$. Here, $\textbf{X}'_{test}$ corresponds to the year $2020$ data, which naturally drifted due to the impact of COVID-19. 
Adding Gaussian noise to other datasets, including COMPAS and GC, led to a slight drop in the ROC AUC and a mild change in the DT feature rankings and their overlap as shown in Table~\ref{tab:results}. This reflected the variation in $\alpha$ or $\mathcal{J}$, and in all cases FMMO triggered an audit alert. In contrast, the addition of uniform noise caused substantial changes in TreeSHAP rankings and a marked drop in $\mathcal{J}$, except for the COMPAS dataset, which showed mixed results. However, in the corresponding experiment with DT on synthetic data, we observed a marked decline in ROC AUC for both ML models, from $94.41$\% to $84.55$\% for RF and from $95.95$\% to $83.97$\% for XGB with Gaussian noise, as shown in Table~\ref{tab:results}. Other metrics, such as DC, CBPE, DT, and TreeSHAP, also indicated substantial performance degradation under covariate drift, prompting FMMO to issue an alert for audit of the ML model. 
These scenarios indicate that local XAI methods such as TreeSHAP may not react to certain feature-level distributional drifts and may not be used alone to identify covariate drifts and should be used along with other statistical methods for drift detection such as DC and global XAI frameworks such as GSM, specifically DT. 

\subsubsection{\textbf{Drift in Sensitive Features}}
\begin{figure*}[ht]
    \centering
    \begin{subfigure}{0.48\linewidth}
        \centering
        \includegraphics[scale=0.35]{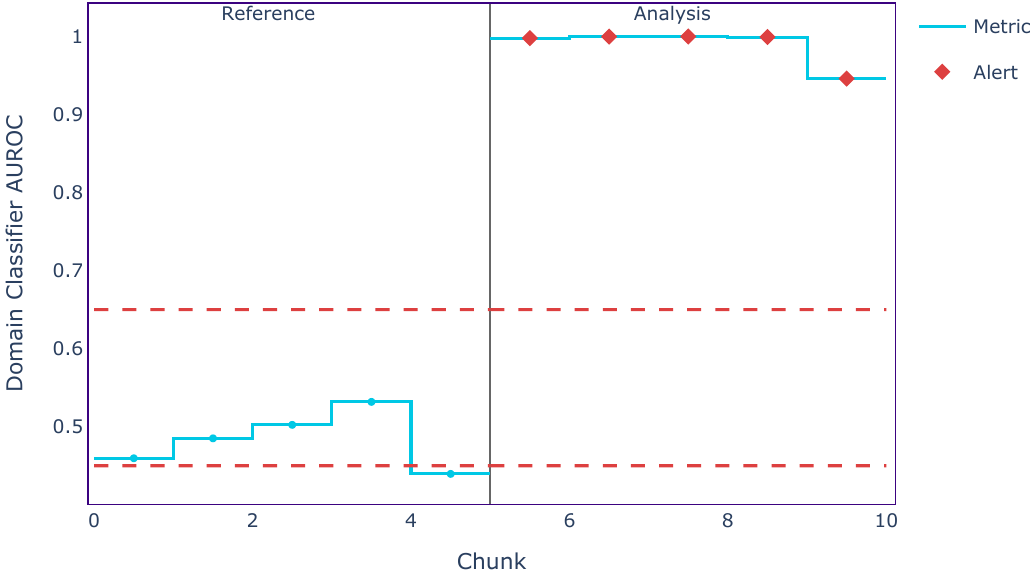}
        \caption{Multivariate drift detection (DC)}
        \label{fig:nml_dcc_compass}
    \end{subfigure}
    \hfill
    \begin{subfigure}{0.48\linewidth}
        \centering
        \includegraphics[scale=0.35]{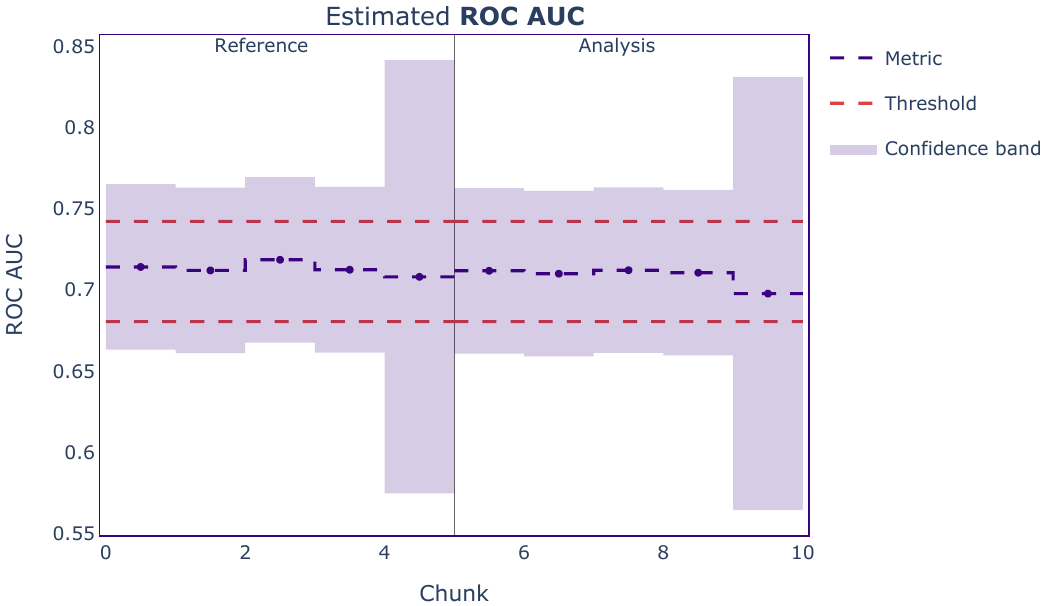}
    \caption{Estimated performance (CBPE)}
    \label{fig:nml_cbpe_compass}
    \end{subfigure}
    
    \begin{subfigure}{0.48\linewidth}
        \centering
        \includegraphics[scale=0.4]{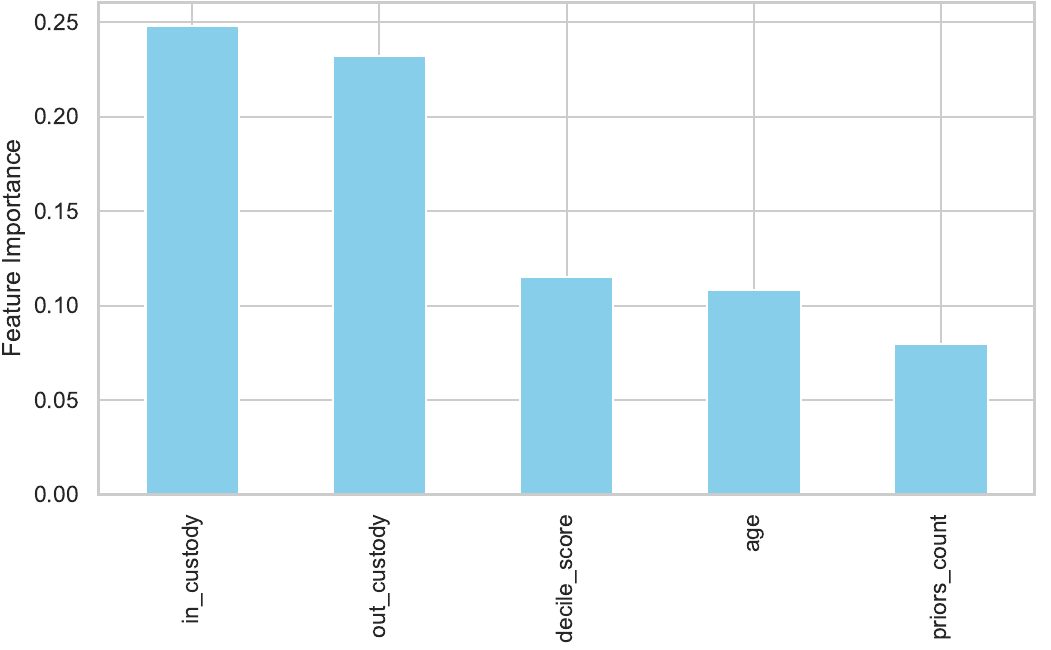}
        \caption{DT - Top 5 features before drift }
        \label{fig:dt_baseline_compas}
    \end{subfigure}
    \hfill
    \begin{subfigure}{0.48\linewidth}
        \centering
        \includegraphics[scale=0.4]{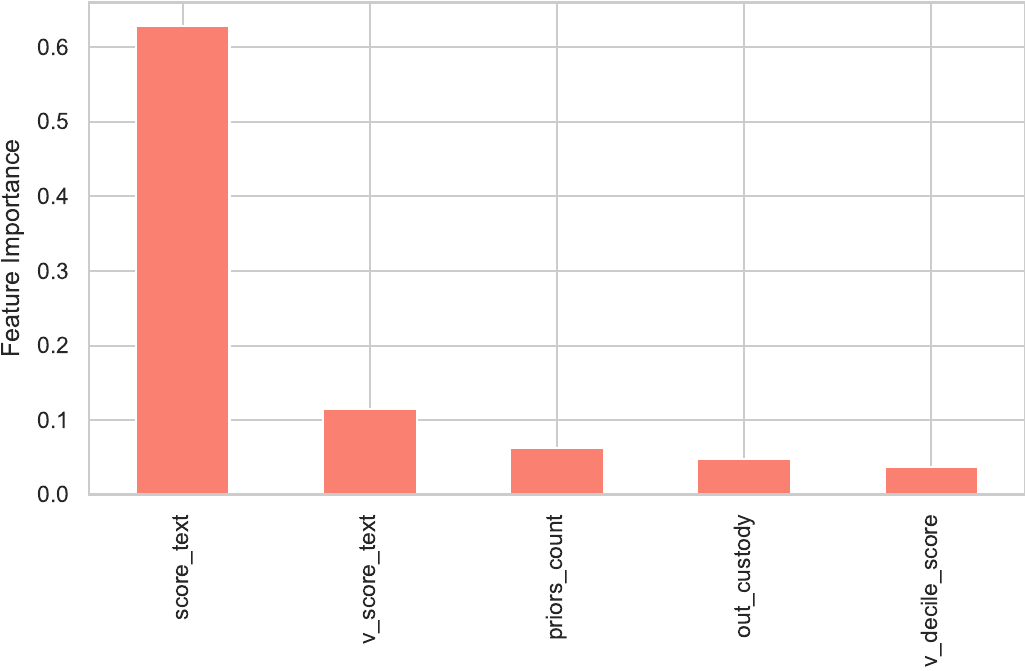}
        \caption{DT - Top 5 features after drift }
        \label{fig:dt_drift_compas}
    \end{subfigure}
    
    \begin{subfigure}{0.48\linewidth}
        \centering
        \includegraphics[scale=0.4]{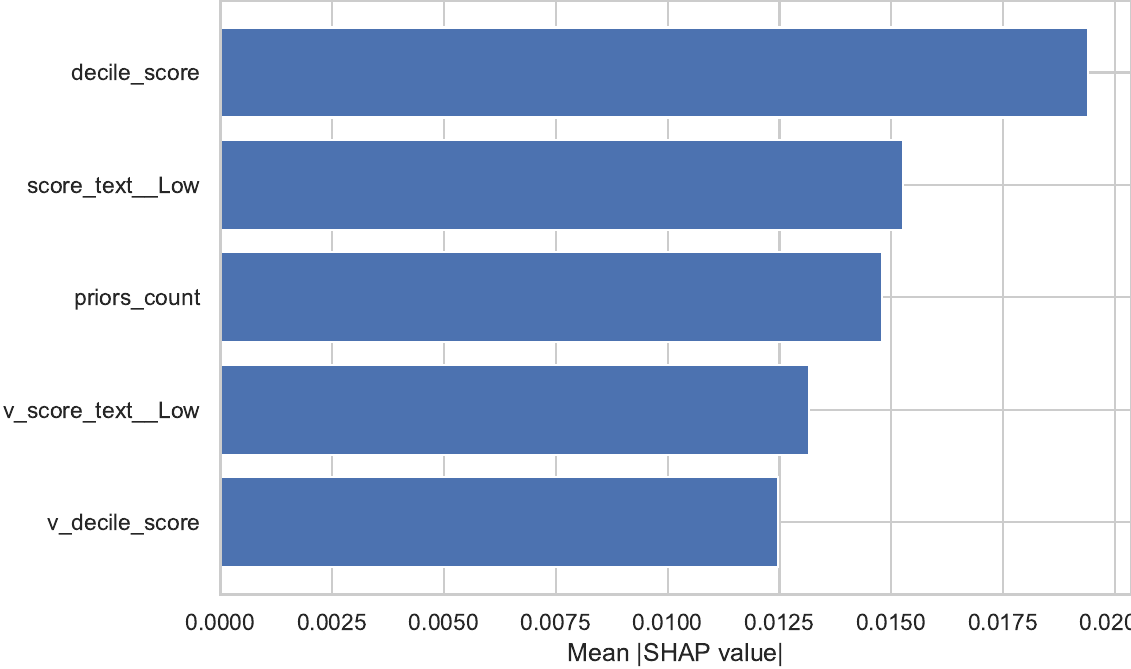}
    \caption{TreeSHAP - Top 5 features before drift}
    \label{fig:dt_shap_baseline_compas}
    \end{subfigure}
    \hfill
    \begin{subfigure}{0.48\linewidth}
        \centering
        \includegraphics[scale=0.4]{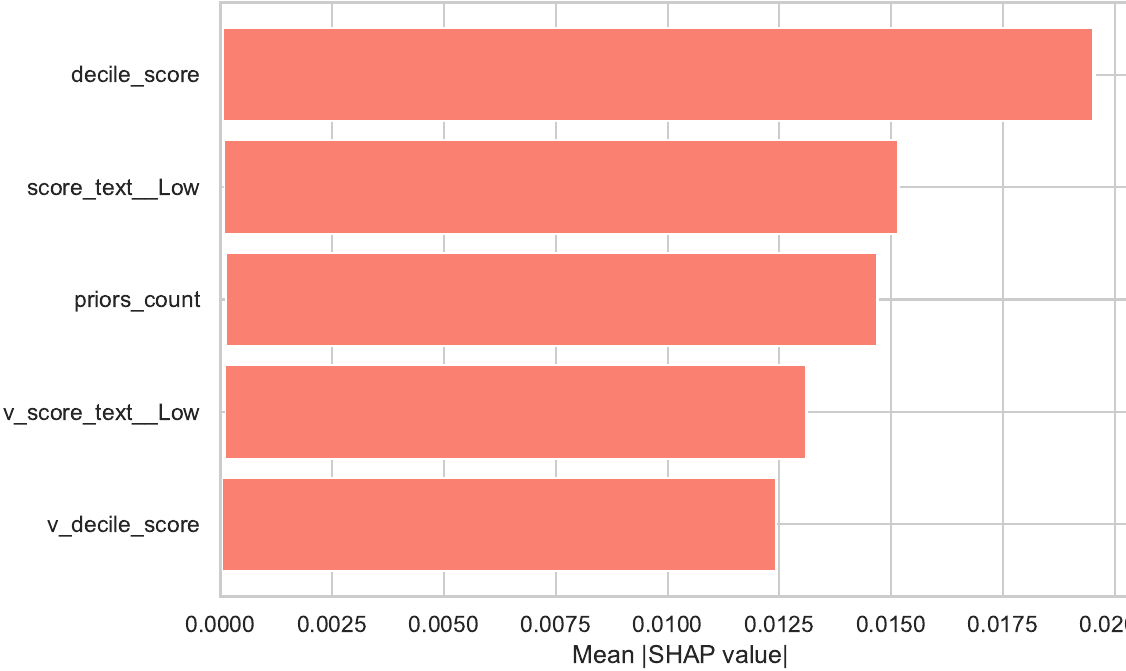}
    \caption{TreeSHAP - Top 5 features after drift}
    \label{fig:dt_shap_drift_compas}
    \end{subfigure}

    \caption{FMMO reports for the COMPAS dataset when Gaussian noise was added}
    \label{fig:nml_compas}
\end{figure*}

When covariate drift is injected into the \enquote{Age} and \enquote{Race} features, overall model accuracy is only marginally affected; however, false positive rates increase disproportionately for the protected group by $32.55$\%, compared to $26$\% increase for the privileged group as shown in Table~\ref{tab:compas_rf_group_metrics}, indicating that drift primarily redistributes error rather than uniformly degrading performance. This fairness deterioration is not revealed by TreeSHAP explanations, as the \enquote{Age} feature does not appear among the top $5$ attributions before or after drift. In contrast, the surrogate DT model ranks \enquote{Age} among the top five influential features under baseline conditions, but this feature disappears from the surrogate explanation after drift as shown in Figures~\ref{fig:dt_baseline_compas} and \ref{fig:dt_drift_compas}. This illustrates how the representation of \enquote{Age} in the explanations impacted under drift, obscuring the source of increased group-level harm.

Crucially, this experiment exposes a dangerous ``fairness blind spot'' in current auditing protocols. While the overall accuracy degradation was minor, the model became significantly more punitive toward the protected group, with the FPR gap widening by over 6\%. An auditor relying solely on SHAP stability (Table 3) would certify this model as compliant, unaware that it has drifted into a discriminatory state. This divergence proves that explanation stability is an insufficient proxy for fairness. FMMO addresses this by coupling the explanation layer with group-specific performance estimation, ensuring that ``silent'' fairness failures are flagged even when the global explanation appears robust.


    

     
    
TreeSHAP and GSM cannot directly expose group-level harm because they describe model behavior rather than disparities in outcomes. Instead, such harm is detected via shifts in FPRs between protected and privileged groups. In our experiments, TreeSHAP explanations remain largely stable even when FPR for the protected group rise substantially, whereas GSM capture structural shifts in feature utilization under drift. Together, these findings show that XAI methods alone cannot reliably detect disparate impacts and need to be combined with group-specific performance monitoring to reveal drift-driven harms.

An important point to highlight is that the FPR is not utilized as a real-time monitoring signal in FMMO. Rather, it is applied retrospectively for experimental evaluation, where it facilitates quantification of the downstream effects of distributional drift that may remain undetected by label-free monitoring signals. In practical deployments where ground-truth labels are not available, the FPR itself cannot be computed directly.

This limitation underscores the need for label-free monitoring approaches in post-deployment machine learning. FMMO addresses a key challenge in post-deployment machine learning by supporting model auditing when ground-truth labels are delayed or unavailable. By combining multiple monitoring signals, it enables earlier identification of deployment conditions that may indicate performance degradation or fairness degradation, thereby supporting more reliable and accountable deployment of ML systems in high-stakes applications.

\subsection{Discussion on Results}
Our first assumption was that if there is a change in the data distribution, then the important features will also change and can be used as a potential drift signal. Our second assumption was that if the important features remain unchanged, the rank will change and can be used as a potential drift signal. During our experiments, we found that most of the time these assumptions failed as shown in Table~\ref{tab:results}, Figure~\ref{fig:dt_shap_baseline_ces} and Figure~\ref{fig:dt_shap_drift_ces}. This is quite possible, as local post-hoc feature attribution methods work on local vicinity and features which are important in local regions may not necessarily be important in the global context. The fundamental design of these local post-hoc feature attribution methods is to explain one instance $\textbf{x} \in \textbf{X}_{test}$ at a time by perturbing data around it and training a simple model (i.e., linear regression) to estimate the decision of black-box ML models by obtaining important features in the local region. Changes in $\textbf{X}_{test}$ may go unnoticed by these local XAI models due to data perturbation. 

This is not the failure of these models; however, it is a limitation because these models are not fundamentally designed  to detect covariate drift. In contrast, the post-hoc global XAI model (surrogate DT) is sensitive to these changes and supports both assumptions in almost all cases, as shown in Table~\ref{tab:results} and can be used to trigger potential covariate drift signals. By evaluating drift indicators based on XAI and model utilization monitoring, our research uncovers significant blind spots in post-hoc XAI tools, emphasizing that consistent explanations do not always guarantee that data have not been altered. The results of our experiments reveal that when data drift occurs, local XAI methods, such as TreeSHAP, often exhibit stability in their attributions, even as model performance changes. However, this stability should not be mistaken for robustness. Local XAI models approximate the behavior of a classifier in a local neighborhood around a single instance, relying on synthetic perturbations of that instance rather than the true global data distribution. Therefore, their explanations remain faithful to the current model function, but may no longer be representative of the intended decision logic or the underlying data semantics. 

In drifted environments, this behavior can create a misleading sense of trustworthiness, as local explanations appear consistent, yet the model may have shifted conceptually. The explanations are correct with respect to the current predictions of the model, but potentially incorrect with respect to the broader context in which these predictions are made. Therefore, the trust that one can place in local explanations should be conditioned on the stability of the surrounding data distribution. Integrating drift detection tools, such as DC or univariate statistical methods, provides an essential context for interpreting these explanations responsibly. A key conclusion of our research is that XAI techniques should be considered as context-specific diagnostics rather than as universal markers for drift detection. 

Another key point is the selection of the values for $\beta$ and $\zeta$. Even without adhering to the strict setting of $\beta = 1$ and $ \zeta = 1$, reducing these parameters to $70$\% still triggers audit alerts under GSM DT, as reported in Table~\ref{tab:results}. The findings in Table~\ref{tab:results} show that audit alerts can occur even when $\mathcal{J}$ and $\alpha$ are well below perfect agreement, in several cases under $70$\%, suggesting that notable drops in performance and fairness can arise even when local explanations seem comparatively stable. 

Although the proposed framework is effective for detecting covariate drift with XAI, it has several limitations, which also suggest directions for future work. For instance, our experiments focused on improving the detection of True Positive (TP) scenarios.  Of course, in doing so, we run the risk of increasing False Positive (FP) and consequently alert fatigue.  Additional negative testing scenarios could help evaluate the risk of False Positive (FP) alerts.

\section{Governance Implications}
The FMMO has direct implications for the governance of deployed ML models, particularly in settings where ground-truth labels are delayed, incomplete, or unavailable. By jointly tracking CBPE and explanation stability, the framework enables ongoing accountability for model behavior beyond initial validation, supporting continuous responsibility for predictive reliability over time. Monitoring explanation stability shows how decision rationales change, helping stakeholders spot cases where models remain accurate but diverge from prior logic. Interpretable monitoring signals also support oversight by providing evidence for model reviews without specialized statistical tests, strengthening accountability and transparency focused governance.

\section{Conclusion \& Future Work}
\label{sec:conclusion}
We present FMMO for flagging models at risk of performance degradation due to covariate drift. The FMMO employs local and global XAI frameworks along with model utilization monitoring with the objective that if any of these observe performance degradation, that could be a potential covariate drift signal, and the framework should raise an alert for the audit of ML models. Through controlled experiments, we demonstrated that post-hoc explanation methods, such as TreeSHAP, tend to remain stable despite measurable performance degradation and substantial shifts in intrinsic global feature dependencies. Although this stability indicates local faithfulness to the model, it is not a reliable indicator of model performance degradation due to covariance drift. In our experiments, covariance drift was reliably identified using FMMO. Our results support the finding that feature attribution in some scenarios can detect covariate drift but exhibits False Negatives (FN), emphasizing that interpretability without being combined with statistical drift can provide misleading explanations. Future work will extend this analysis by incorporating multi-model comparisons, per-feature drift correlation, and uncertainty estimation to advance a more integrated framework for reliable detection of other types of drift.  



\end{document}